\documentclass{article}

\usepackage{xcolor}
\definecolor{mydarkblue}{rgb}{0,0.15,0.7}
\usepackage[colorlinks, urlcolor=mydarkblue, citecolor=mydarkblue, linkcolor=mydarkblue]{hyperref}

\usepackage[final, nonatbib]{midl_2018}

\usepackage[utf8]{inputenc} 
\usepackage[T1]{fontenc}    
\usepackage{hyperref}       
\usepackage{url}            
\usepackage{booktabs}       
\usepackage{amsfonts}       
\usepackage{nicefrac}       
\usepackage{microtype}      
\usepackage{amsmath}
\usepackage{algorithmic}
\usepackage{algorithm}
\usepackage{graphicx}
\usepackage{subcaption}
\usepackage{siunitx}
\usepackage{textcomp}
\usepackage{amssymb}
\usepackage{multirow}
\usepackage{threeparttable}

\title{Weakly Supervised Lesion Localization With Probabilistic-CAM Pooling}

\author{
  Wenwu Ye \\
  JF Healthcare \\
  \texttt{wenwu.ye@jfhealthcare.com} \\
  \And
  Jin Yao \\
  JF Healthcare \\
  \texttt{jin.yao@jfhealthcare.com} \\
  \And
  Hui Xue \\
  JF Healthcare \\
  \texttt{hui.xue@jfhealthcare.com} \\
  \And
  Yi Li \\
  Greybird Ventures LLC \\
  \texttt{yil8@uci.edu} \\
}

\begin{document}

\maketitle

\begin{abstract}
Localizing thoracic diseases on chest X-ray plays a critical role in clinical
practices such as diagnosis and treatment planning. However, current deep learning
based approaches often require strong supervision, e.g. annotated bounding boxes,
for training such systems, which is infeasible to harvest in large-scale. We present
Probabilistic Class Activation Map (PCAM) pooling, a novel global pooling operation for lesion
localization with only image-level supervision. PCAM pooling explicitly leverages
the excellent localization ability of CAM~\cite{zhou2016learning}
during training in a probabilistic fashion. Experiments on the ChestX-ray14~\cite{wang2017chestx}
dataset show a ResNet-34~\cite{he2016deep} model trained with PCAM pooling
outperforms state-of-the-art baselines on both the classification task and the
localization task. Visual examination on the probability maps generated by PCAM
pooling shows clear and sharp boundaries around lesion regions compared to the
localization heatmaps generated by CAM. PCAM pooling is open sourced at \url{https://github.com/jfhealthcare/Chexpert}.

\end{abstract}

\section{Introduction}
Computer-aided thoracic disease diagnosis based on chest X-ray has been
significantly advanced by deep convolutional neural networks (DCNNs) in recent
years~\cite{wang2017chestx,tang2018attention,sedai2018deep,wang2018chestnet}. 
Most of these approaches are formulated as a multi-task binary classification
problem, where a CNN is trained to predict the risk probabilities of different
thoracic diseases. In clinical practices, visual localization of the
lesions on chest X-ray, such as heatmaps or segmentation masks,
is also preferred to provide interpretable supports for the classification results.
Precise lesion localization often requires training CNNs with strong supervision,
such as bounding boxes~\cite{wang2017chestx}, beyond merely image-level labels.
However, accurately annotating lesion locations is difficult, time-consuming, and
infeasible to practice in large-scale. For example, one of the largest
publicly available chest X-ray datasets ChestX-ray14~\cite{wang2017chestx} contains
more than one hundred thousands images with image-level labels, among which
only less than one thousand images are further annotated with bounding boxes for
benchmarking. Therefore, weakly supervised lesion localization on chest X-ray based
on image-level labels remains a challenge but vital problem for computer-aided
thoracic disease diagnosis.

The recent work of Class Activation Map (CAM)~\cite{zhou2016learning} demonstrates
the excellent localization ability of CNNs trained on nature images with only image-level
supervision. On chest X-ray images, CAM and its variations
have also been used for lesions localization~\cite{wang2017chestx,tang2018attention,sedai2018deep,wang2018chestnet}.
However, most of these approaches utilize CAM as a post-processing technique to
first generate lesion localization heatmaps, then threshold the heatmap scores
and generate lesion bounding boxes. We argue that it may be beneficial to leverage
CAM even during training given its excellent localization ability.

\begin{figure}[t]
   \begin{center}
      \includegraphics[width=0.95\linewidth]{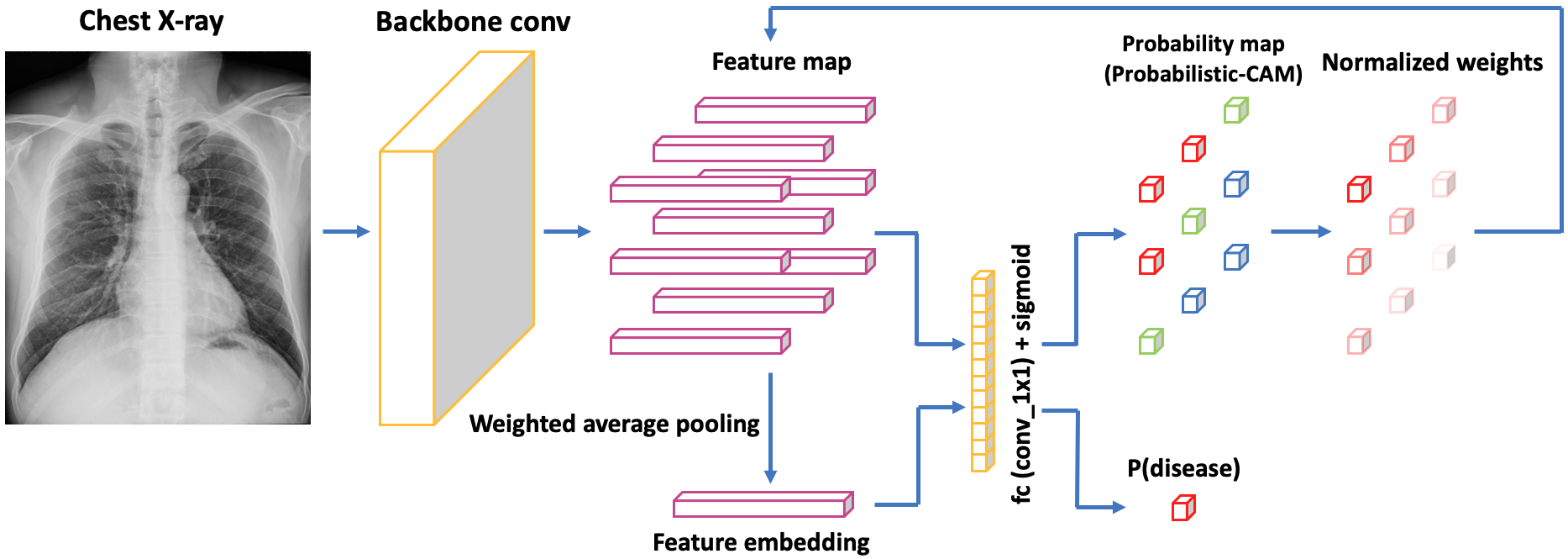}
   \end{center}
      \caption{The framework of Probabilistic-CAM (PCAM) pooling.}
   \label{fig1}
\end{figure}

In this work, we propose a novel and simple extension to the CAM-based framework for lesion
localization on chest X-ray with image-level supervision. Specifically, we propose
a new global pooling operation that explicitly leverages CAM for localization during
training in a probabilistic fashion, namely Probabilistic-CAM (PCAM) pooling. Figure~\ref{fig1}
shows the framework of PCAM pooling. A fully convolutional backbone network first
processes the input chest
X-ray image and generates a feature map. Then, for a particular label of thoracic
disease, e.g. ``Pneumonia'', each feature embedding within the feature map goes through a fully
connected (fc) layer implemented as a $1 \times 1$ convolutional layer and generates
the class activation score that monotonically measure the disease likelihood
of each embedding. Unlike the standard practice that directly uses the class activation score for localization, we
further bound it with the sigmoid function and interpret the output as
the disease probability of each embedding. Finally, the output probability
map is normalized to attention weights of each embedding, following the
multiple-instance learning (MIL) framework~\cite{ilse2018attention,yao2018weakly},
which are used to pool the original feature map by weighted average pooling.
The pooled embedding goes through
the same fc layer introduced above and generates the image-level disease probability
for training. During inference time, we directly use the probability map for lesion
localization, and apply a simple probability thresholding to obtain disease regions
and bounding boxes. 

PCAM pooling does not introduce any additional training parameters and is easy
to implement. We experiment the efficacy of PCAM pooling for lesion localization
with image-level supervision on the ChestX-ray14~\cite{wang2017chestx} dataset. A
ResNet-34~\cite{he2016deep} model trained with PCAM pooling significantly outperforms the ChestX-ray14
baseline\cite{wang2017chestx} in both the classification task and the localization
task. Qualitative visual examination shows the probability maps generated by PCAM
pooling tend to have clear and sharp boundaries around lesion regions compared
to the typical class activation map.

\section{Related work}
\subsection{Weakly supervised lesion localization}
Various methods have been proposed to localize lesions on chest X-ray through
image-level supervision~\cite{wang2017chestx,tang2018attention,sedai2018deep,yao2018weakly,wang2018chestnet}.
Wang et al.~\cite{wang2017chestx} introduce the ChestX-ray14 dataset, together
with a baseline for evaluating weakly supervised lesion
localization. Instead of the typical global average pooling, Wang et al. use
the Log-Sum-Exp (LSE) pooling~\cite{pinheiro2015image} to encourage model training
focusing more on the discriminative regions of the chest X-ray. Sedai et al.~\cite{sedai2018deep}
utilize the intermediate feature maps from CNN layers at different scales together
with learned layer reference weights to improve the localization performance of
small lesions. Tang et al.~\cite{tang2018attention}
combine CNN with attention-guided curriculum learning to gradually learn distinctive
convolutional features for localization. Most of these
approaches utilize the standard CAM technique to localize lesions, where our
proposed PCAM pooling serves as an extension to the standard CAM to improve lesion localization
with image-level supervision.

\subsection{Global pooling}\label{sec2.2}
Global average pooling is arguably the most widely used global pooling operation
for image classification. While it is less prone to overfitting as the network
tries to identify all discriminative regions of an object in natural images~\cite{zhou2016learning},
it may also fail to highlight the most discriminative regions within a chest X-ray,
given most chest X-ray images share the same anatomic structures and may only differ
in fine-grained lesion regions. Therefore, different global pooling operations
have also been used to analyze chest X-rays. For example, Wang et al.~\cite{wang2017chestx} uses LSE pooling,
which can be viewed as an adjustable operation between max pooling and average
pooling by controlling the hyper-parameter $\gamma$.

We summarize the mathematical differences and correlations between different
global pooling operations in Table~\ref{tab1}. Particularly, $X$ with shape
$(C, H, W)$ denotes the feature map from the last convolutional layer of a CNN.
$x$ denotes the feature embedding of length $C$ after global pooling. $C$ denotes
the channel dimension, $H$ and $W$ denote the height and width of the feature map.

\begin{table}
   \centering
   \caption{Different types of global pooling.}\label{tab1}
   \begin{tabular}{|l|l|}
   \hline
   Pooling type &  Formulation\\
   \hline
   Average       &  $x_c = \sum_{i,j}^{H,W} w_{c,i,j} X_{c,i,j} ,\ w_{c,i,j} = \frac{1}{H*W}$  \\
   Linear~\cite{wang2019comparison}        &  $x_c = \sum_{i,j}^{H,W} w_{c,i,j} X_{c,i,j} ,\ w_{c,i,j} = \frac{X_{c,i,j}}{\sum_{i,j}^{H,W} X_{c,i,j}}$  \\
   Exponential~\cite{wang2019comparison}    &  $x_c = \sum_{i,j}^{H,W} w_{c,i,j} X_{c,i,j} ,\ w_{c,i,j} = \frac{\exp(X_{c,i,j})}{\sum_{i,j}^{H,W} \exp(X_{c,i,j})}$  \\
   LSE~\cite{pinheiro2015image}   &  $x_c = \frac{1}{\gamma} \log\left[ \frac{1}{H*W} \sum_{i,j}^{H,W} \exp(\gamma X_{c,i,j}) \right]$ \\
   LSE-LBA~\cite{yao2018weakly}   &  $s = \frac{1}{\gamma_0 + \exp(\beta)} \log\left[ \frac{1}{H*W} \sum_{i,j}^{H,W} \exp\left[(\gamma_0 + \exp(\beta)) S_{i,j}\right] \right]$ \\
   Attention~\cite{ilse2018attention}     &  $x = \sum_{i,j}^{H,W} w_{i,j} X_{i,j} ,\ w_{i,j} = \frac{\exp\left[\mathbf{w}^\intercal\tanh(\mathbf{V} X_{i,j})\right]}{\sum_{i,j}^{H,W} \exp\left[\mathbf{w}^\intercal\tanh(\mathbf{V} X_{i,j})\right]} $ \\
   \hline
   \end{tabular}
\end{table}

We can see that, global pooling operations differ mainly in the ways to compute the
weights for feature map averaging. Note that, most of the pooling is performed for each
channel $X_{c,i,j}$ independently, except for Attention pooling~\cite{ilse2018attention} that
computes the attention weight at embedding level $X_{i,j}$ with extra trainable parameters,
i.e. $\mathbf{V}, \mathbf{w}$ in Table~\ref{tab1}. All the channels within the
same embedding share the same weight. The basic assumption of Attention
pooling follows the multiple-instance learning (MIL) framework, that treats each
embedding as an instance, and the chest X-ray as a bag of instances is positive,
e.g. certain thoracic disease, as long as one of the instance is positive.
PCAM pooling follows the same MIL framework, but uses a different
method to compute the attention weight for each embedding based on
the localization ability of CAM without introducing extra trainable parameters.
LSE-LBA~\cite{yao2018weakly} also falls within the MIL framework, but it performs
the pooling operation on the saliency map $S$ of shape $(H, W)$ instead of the
feature map $X$ to obtain a saliency score $s$ for training.

\section{Probabilistic-CAM (PCAM) Pooling}
The main idea of PCAM pooling is to explicitly leverage the localization ability
of CAM~\cite{zhou2016learning} through the global pooling operation during training.
Using the same notation from Section~\ref{sec2.2}, given a fully convolutional
network trained for multi-task binary classification, the class activation map of
a particular thoracic disease is given by $\{s_{i,j} = \mathbf{w}^\intercal X_{i,j} + b|i,j \in H,W\}$.
$X_{i,j}$ is the feature embedding of length $C$ at the position $(i, j)$
of a feature map $X$ with shape $(C, H, W)$ from the last convolutional layer.
$\mathbf{w}, b$ are the weights and bias of the last fc layer for binary
classification. In other words, $s_{i,j}$ is the logit before sigmoid function
under the binary classification setting. $s_{i,j}$ monotonically measures the
disease likelihood of $X_{i,j}$, and is used to generate the localization heatmap
after the model is trained in the standard CAM framework.

PCAM pooling utilizes $s_{i,j}$ to guide lesion localization during training
through the global pooling operation under the MIL framework~\cite{ilse2018attention}.
The MIL framework assumes the chest X-ray as a bag of embeddings is positive, as
long as one of the embedding is positive. To measure each embedding's contribution
to the whole bag, the MIL framework assigns normalized attention weights to each embedding for
weighted global average pooling~\cite{ilse2018attention}. Because the numerical
range of $s_{i,j}$ is unbounded, i.e. $s_{i,j} \in (-\infty, +\infty)$ in theory,
it's neither interpretable nor directly applicable to compute the attention weights.
Therefore, we further bound $s_{i,j}$ with the sigmoid function, $p_{i,j} = \text{sigmoid}(s_{i,j})$,
and normalize $p_{i,j}$ as the attention weights. In summary, PCAM
pooling can be formulated as
\begin{eqnarray}\label{eq1}
x = \sum_{i,j}^{H,W} w_{i,j} X_{i,j} ,\ w_{i,j} = \frac{\text{sigmoid}(\mathbf{w}^\intercal X_{i,j} + b)}{\sum_{i,j}^{H,W} \text{sigmoid}(\mathbf{w}^\intercal X_{i,j} + b)}
\end{eqnarray}
where $w_{i,j}$ is the attention weight for $X_{i,j}$ and $x$ is the pooled
feature embedding which goes through the same fc layer for final image level classification.

During the inference time, we interpret the sigmoid-bounded $p_{i,j}$ as the
probability of embedding $X_{i,j}$ being positive, thus named Probabilistic-CAM,
and we use the probability map $\{p_{i,j}|i,j \in H,W\}$ as the localization heatmap.
We use a simple probability thresholding on the probability map to obtain regions
of interest. In comparison, because $s_{i,j}$ is unbounded with different numerical
ranges in different chest X-ray images, it is usually normalized into $[0, 255]$
within each image and thresholded with some ad-hoc ranges, e.g. $[60, 180]$
in~\cite{wang2017chestx}, to generate regions of interest. We show in Section~\ref{sec4.3}
section that PCAM pooling generates localization heatmaps with better visual quality
around lesion boundary regions compared to the standard CAM.

\section{Experiments}
\subsection{Dataset and experiments setup}\label{sec4.1}
We evaluate lesion localization with image-level supervision on the ChestX-ray14~\cite{wang2017chestx}
dataset, which contains 112,120 frontal-view chest X-ray images with 14 thoracic
disease labels. 8 out of the 14 diseases are further annotated with 984 bounding
boxes. We randomly split the official train\_valid set into $75\%$ for
training and $25\%$ for validation. On the official test set, we evaluate the
classification task on the 14 diseases and the localization task on the 8 diseases
that have bounding boxes. Note that the 984 bounding boxes are not used for training.


We use ResNet-34~\cite{he2016deep} as the backbone network and process
the input images on the original $1024 \times 1024$ scale following~\cite{wang2017chestx}.
The network is trained with a batch size of 36 and a learning rate of $1e^{-4}$
for 10 epoches. We balance the binary cross entropy loss of positive and negative
samples within each batch following~\cite{wang2017chestx}. For the localization
task, we first apply a  probability of 0.9 to threshold the probability maps
from PCAM pooling, then generate bounding boxes that cover the isolated regions
in the binary masks. To compare the visual quality of localization heatmaps from
PCAM pooling with previous methods, we also train a ResNet-34 with LSE pooling
following~\cite{wang2017chestx}. We normalize the class activation maps from LSE
pooling into $[0, 255]$ for each image individually, and then apply a threshold
of 180 following~\cite{wang2017chestx}. Note that the performances of LSE pooling
reported in Table~\ref{tab2} and Table~\ref{tab3} are from Wang et al.~\cite{wang2017chestx}.

\subsection{Classification task}\label{sec4.2}

\begin{table}
\centering
\caption{Classification AUCs of the 14 diseases on the ChestX-ray14 test set.}\label{tab2}
\begin{tabular}{|l|c|c|c|c|c|}
\hline
Method             & LSE~\cite{wang2017chestx} & LSE-LBA~\cite{yao2018weakly} & AGCL~\cite{tang2018attention} & ChestNet~\cite{wang2018chestnet} & PCAM pooling \\
\hline
Atelectasis        & 0.700 & 0.733 & 0.756 & 0.743 & \textbf{0.772} \\
Cardiomegaly       & 0.810 & 0.856 & \textbf{0.887} & 0.875 & 0.864 \\ 
Effusion           & 0.759 & 0.806 & 0.819 & 0.811 & \textbf{0.825} \\
Infiltration       & 0.661 & 0.673 & 0.689 & 0.677 & \textbf{0.694} \\
Mass               & 0.693 & 0.777 & \textbf{0.814} & 0.783 & 0.813 \\
Nodule             & 0.669 & 0.718 & 0.755 & 0.698 & \textbf{0.783} \\
Pneumonia          & 0.658 & 0.684 & \textbf{0.729} & 0.696 & 0.721 \\
Pneumothorax       & 0.799 & 0.805 & 0.850 & 0.810 & \textbf{0.868} \\
Consolidation      & 0.703 & 0.711 & 0.728 & 0.726 & \textbf{0.732} \\
Edema              & 0.805 & 0.806 & \textbf{0.848} & 0.833 & 0.833 \\
Emphysema          & 0.833 & 0.842 & 0.908 & 0.822 & \textbf{0.931} \\
Fibrosis           & 0.786 & 0.743 & 0.818 & 0.804 & \textbf{0.819} \\
Pleural Thickening & 0.684 & 0.724 & 0.765 & 0.751 & \textbf{0.788} \\
Hernia             & 0.872 & 0.775 & 0.875 & \textbf{0.900} & 0.784 \\             
\hline
\end{tabular}
\end{table}

Table~\ref{tab2} shows the Area Under the receiver operating characteristic Curves
(AUCs) of the classification task on the 14 thoracic diseases from the ChestX-ray14
official test set. PCAM pooling
outperforms most of the other state-of-the-art methods including the baseline
reported in ChestX-ray14~\cite{wang2017chestx}. We suspect explicitly utilizing CAM for localization
during training may also benefit the classification task. Note the results from
AGCL~\cite{tang2018attention} are obtained by training with additional severity-level
information.

\subsection{Localization task}\label{sec4.3}

\begin{table}
\centering
\caption{Localization accuracies and average false positives of the 8 diseases that have bounding boxes on the ChestX-ray14 test set.}\label{tab3}
\begin{tabular}{|l|c|c|c|c|c|c|c|c|}
\hline
Method                              & AT     & CM     & EF     & IF     & MS    & ND    & PMN   & PMT \\
\hline
\multicolumn{9}{|c|}{Localization accuracy, IoBB $>$ 0.5} \\
\hline
LSE-baseline~\cite{wang2017chestx}  & 0.2833 & 0.8767 & 0.3333 & 0.4227 & 0.1411 & 0.0126 & 0.3833 & 0.1836 \\
PCAM pooling                        & \textbf{0.3500} & \textbf{0.9657} & \textbf{0.5359} & \textbf{0.7642} & \textbf{0.4118} & \textbf{0.0759} & \textbf{0.7667} & \textbf{0.1939} \\
\hline
\multicolumn{9}{|c|}{Average false positive} \\
\hline
LSE-baseline~\cite{wang2017chestx}  & 1.020 & 0.563 & \textbf{0.927} & \textbf{0.659} & \textbf{0.694} & \textbf{0.619} & \textbf{1.013} & \textbf{0.529} \\
PCAM pooling                        & \textbf{0.867} & \textbf{0.021} & 1.137 & 1.805 & 1.000 & 1.228 & 1.200 & 1.684 \\
\hline
\end{tabular}
    \begin{tablenotes}
      \small
      \item AT: Atelectasis, CM: Cardiomegaly, EF: Effusion, IF: Infiltration,
      MS: Mass, ND: Nodule, PMN: Pneumonia, PMT: Pneumothorax.
    \end{tablenotes}
\end{table}

Table~\ref{tab3} shows the localization accuracies and the average false positives
of the localization task on the 8 thoracic diseases that have bounding boxes. We use Intersection over
the predicted B-Box area ratio (IoBB) to measure the overlap between predicted
bounding boxes and ground truth bounding boxes annotated by radiologists
following~\cite{wang2017chestx}. A correct localization is defined as at least
one predicted bounding box is overlapped with the ground truth bounding box with
IoBB $>$ 0.5~\cite{wang2017chestx}. PCAM pooling outperforms the baseline
localization accuracy~\cite{wang2017chestx} by a significant margin on all of
the diseases, demonstrating its efficacy in weakly supervised lesion localization.

\begin{figure}
    \centering
    \begin{subfigure}[b]{0.48\linewidth}
        \includegraphics[width=\linewidth]{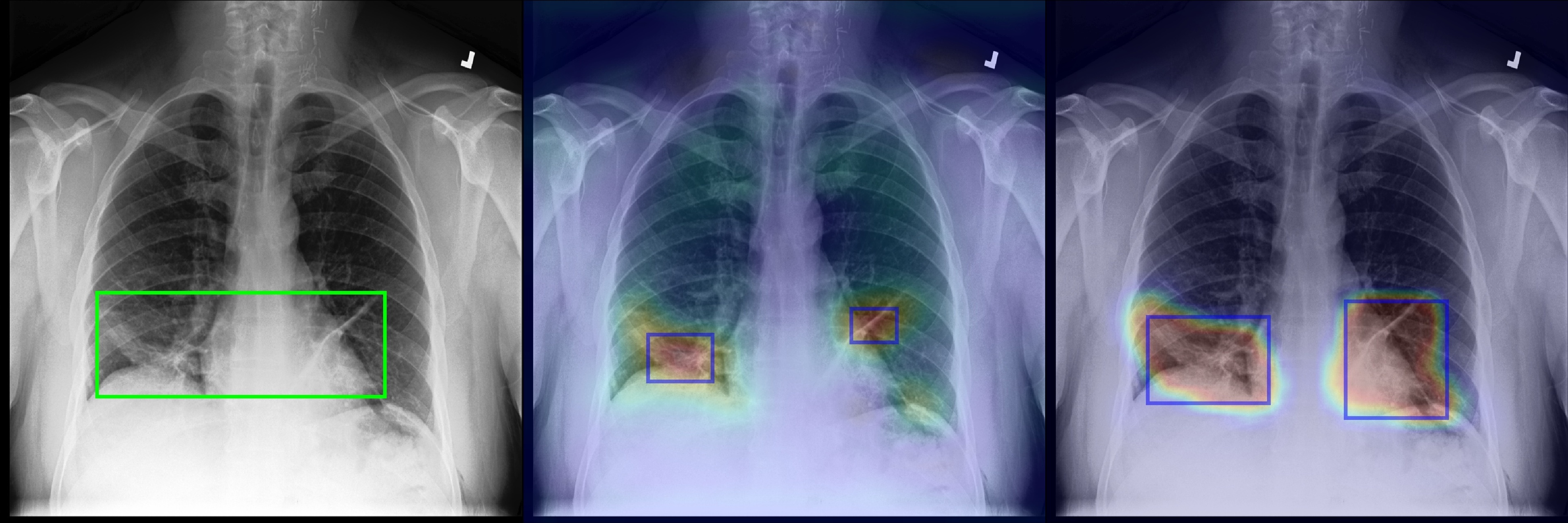}
        \caption{Atelectasis}
        \label{fig2:a}
    \end{subfigure}
    ~
    \begin{subfigure}[b]{0.48\linewidth}
        \includegraphics[width=\linewidth]{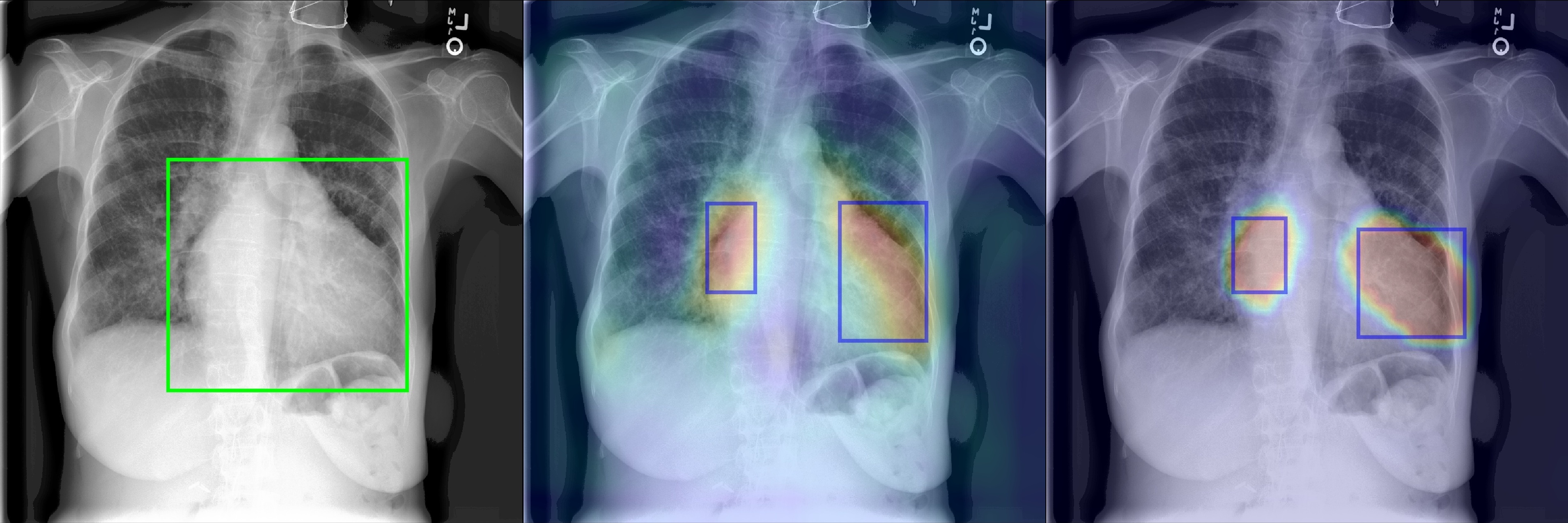}
        \caption{Cardiomegaly}
        \label{fig2:b}
    \end{subfigure}
    ~
    \begin{subfigure}[b]{0.48\linewidth}
        \includegraphics[width=\linewidth]{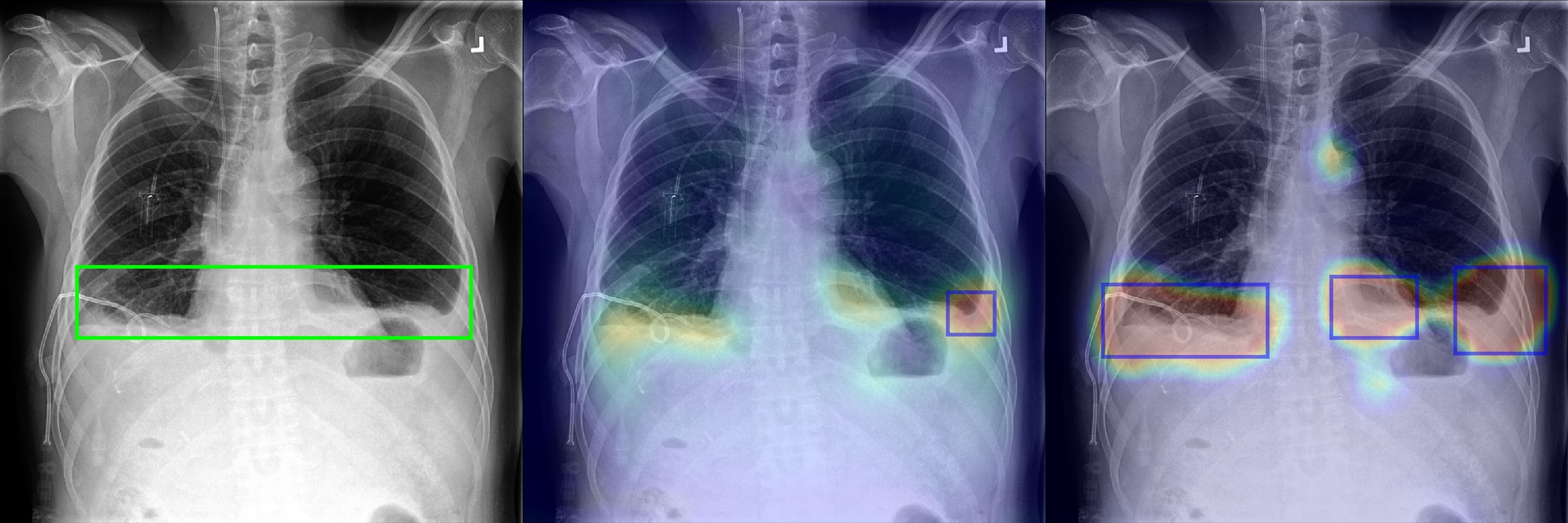}
        \caption{Effusion}
        \label{fig2:c}
    \end{subfigure}
    ~
    \begin{subfigure}[b]{0.48\linewidth}
        \includegraphics[width=\linewidth]{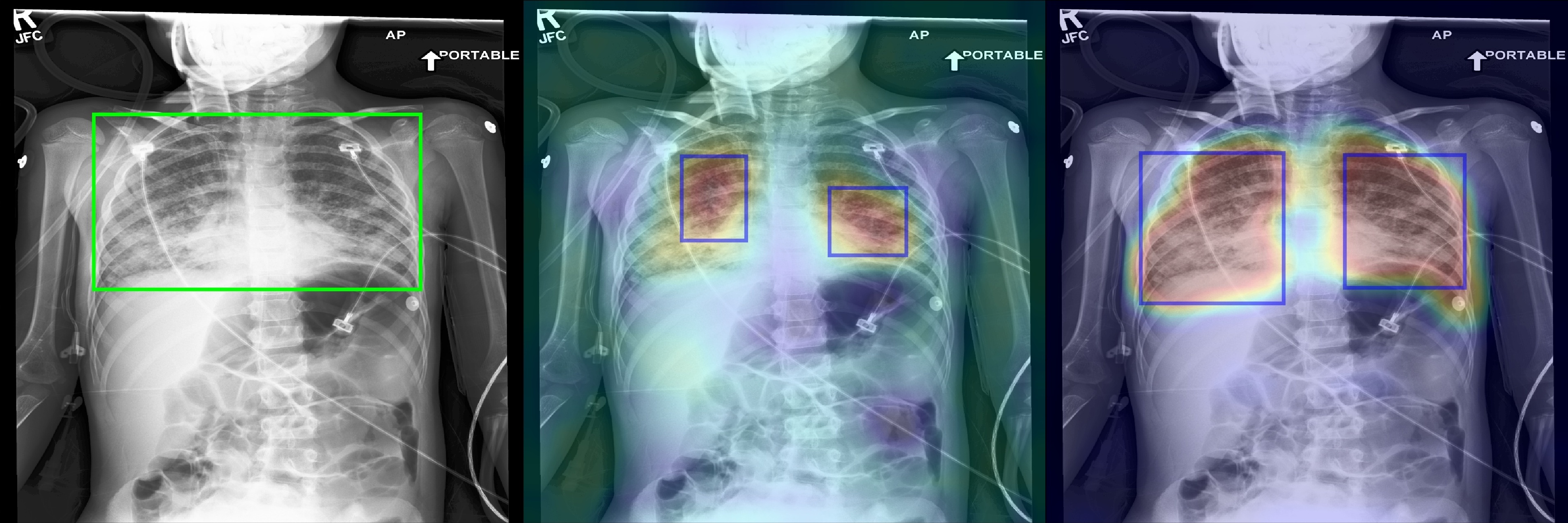}
        \caption{Pneumonia}
        \label{fig2:d}
    \end{subfigure}
    ~
    \begin{subfigure}[b]{0.48\linewidth}
        \includegraphics[width=\linewidth]{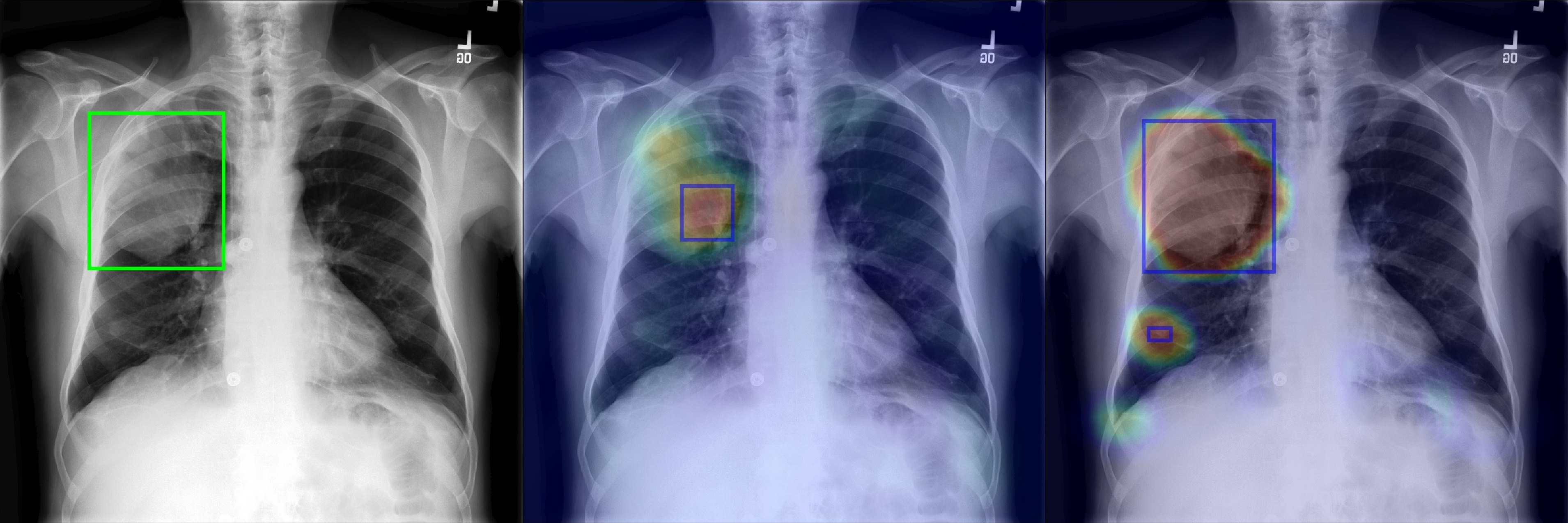}
        \caption{Mass}
        \label{fig2:e}
    \end{subfigure}
    ~
    \begin{subfigure}[b]{0.48\linewidth}
        \includegraphics[width=\linewidth]{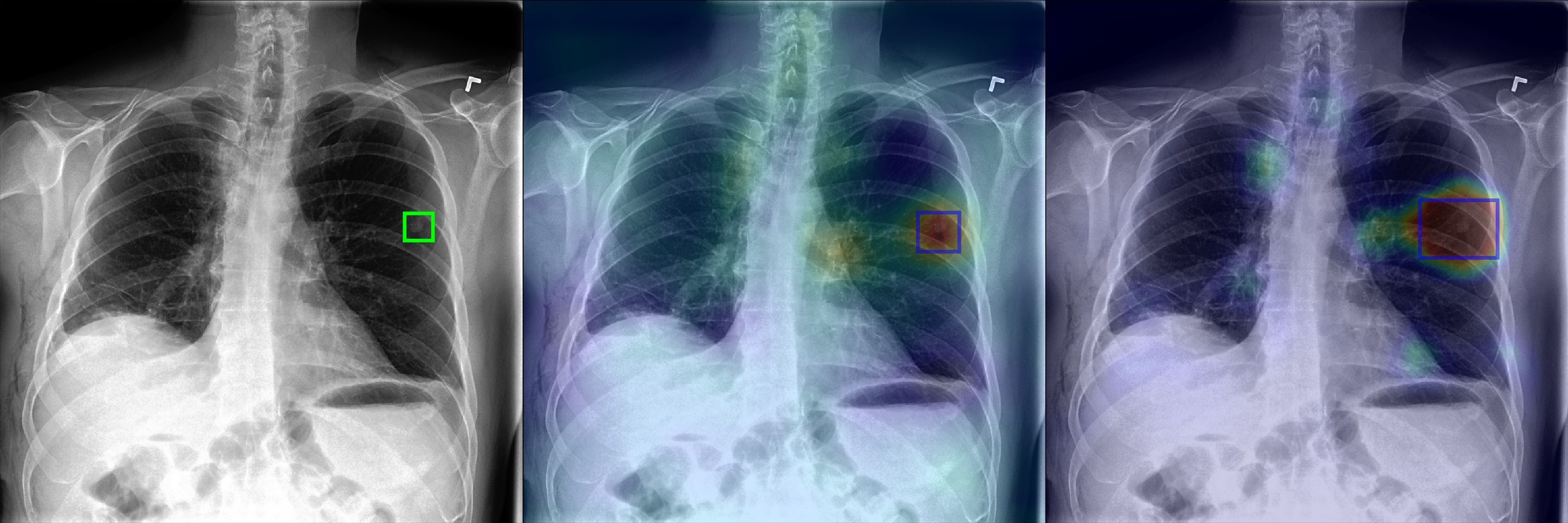}
        \caption{Nodule}
        \label{fig2:f}
    \end{subfigure}
    ~
    \caption{Selected samples of localization heatmaps and their bounding boxes
    generated by LSE pooling and PCAM pooling on the test set of
    ChestX-ray14~\cite{wang2017chestx}. In each subfigure, the left panel is the original
    chest X-ray with the ground truth bounding boxes (green), the middle panel
    is the class activation map and predicted bounding boxes (blue) by LSE pooling,
    the right panel is the probability map and predicted bounding boxes (blue) by
    PCAM pooling.}
    \label{fig2}
\end{figure}

Figure~\ref{fig2} shows a few selected examples of the probability maps generated
by PCAM pooling and the class activation maps generated by LSE pooling together with
the predicted bounding boxes. Compared to the class activation maps, the probability
maps are visually more clear with sharp boundaries around lesion regions.
We attribute the improved visual quality to the probabilistic interpretation of
the sigmoid-bounded class activation map and explicitly using it for training
with global pooling.

We notice the probability maps generated by PCAM pooling tend to enlarge regions
of interest in general than class activation maps from LSE pooling, especially when
the ground truth regions are small, such as ``Nodule'' in Figure~\ref{fig2}.
This may explain the fact that PCAM pooling has relatively larger average false
positives than CAM with LSE pooling.

\section{Conclusion}
In this work, we present Probabilistic-CAM (PCAM) pooling, a new global pooling
operation to explicitly leverage the localization ability of CAM for training.
PCAM pooling is easy to implement and does not introduce any additional training
parameters. Experiments of weakly supervised lesion localization on the ChestX-ray14~\cite{wang2017chestx}
dataset demonstrate its efficacy in improving both the classification task and
the localization task compared to several state-of-the-art baselines. Visual
inspection of the probability maps generated by PCAM pooling shows clear and sharp
boundaries around lesion regions compared to the standard class activation maps.

Currently, PCAM pooling tends to generate localization maps that enlarge regions
of interest, which may increase false positives especially for small lesions. We
are working on reducing this effect as our future direction.


\small

\bibliographystyle{abbrv}

\end{document}